\documentclass[12pt]{spieman}  
\usepackage{amsmath,amsfonts,amssymb}
\usepackage{graphicx}
\usepackage{epstopdf} 
\usepackage{setspace}
\usepackage{tocloft}
\usepackage{lineno}
\usepackage{soul, color, xcolor}
\soulregister{\cite}7 
\soulregister{\citep}7 
\soulregister{\citet}7
\soulregister{\ref}7
\soulregister{\pageref}7 
\soulregister{\subsubsection}7
\usepackage[utf8]{inputenc}

\title{Enhancing breast cancer detection on screening mammogram using self-supervised learning and a hybrid deep model of Swin Transformer and CNN}

\author[a,b]{Han Chen}
\author[a,b]{Anne L. Martel}
\affil[a]{Physical Sciences, Sunnybrook Research Institute, Toronto, ON, Canada}
\affil[b]{Medical Biophysics, University of Toronto, Toronto, ON, Canada}

\cftpagenumbersoff{figure}
\cftpagenumbersoff{table} 
\begin{document} 
\maketitle

\begin{abstract}
\\
\textbf{Purpose:}
The scarcity of high-quality curated labeled medical training data remains one of the major limitations in applying artificial intelligence (AI) systems to breast cancer diagnosis. Deep models for mammogram analysis and mass (or micro-calcification) detection require training with a large volume of labeled images, which are often expensive and time-consuming to collect. To reduce this challenge, we proposed a novel method that leverages self-supervised learning (SSL) and a deep hybrid model, named \textbf{HybMNet}, which combines local self-attention and fine-grained feature extraction to enhance breast cancer detection on screening mammograms.
\\
\textbf{Approach:}
Our method employs a two-stage learning process: (1) SSL Pretraining: We utilize EsViT, a SSL technique, to pretrain a Swin Transformer (Swin-T) using a limited set of mammograms. The pretrained Swin-T then serves as the backbone for the downstream task. (2) Downstream Training: The proposed HybMNet combines the Swin-T backbone with a CNN-based network and a novel fusion strategy. The Swin-T employs local self-attention to identify informative patch regions from the high-resolution mammogram, while the CNN-based network extracts fine-grained local features from the selected patches. A fusion module then integrates global and local information from both networks to generate robust predictions. The HybMNet is trained end-to-end, with the loss function combining the outputs of the Swin-T and CNN modules to optimize feature extraction and classification performance.
\\
\textbf{Results:}
The proposed method was evaluated for its ability to detect breast cancer by distinguishing between benign (normal) and malignant mammograms. Leveraging SSL pretraining and the HybMNet model, it achieved AUC of 0.864 (95\% CI: 0.852, 0.875) on the CMMD dataset and 0.889 (95\% CI: 0.875, 0.903) on the INbreast dataset, highlighting its effectiveness.
\\
\textbf{Conclusions:}
The quantitative results highlight the effectiveness of our proposed HybMNet and the SSL pretraining approach. Additionally, visualizations of the selected ROI patches show the model's potential for weakly-supervised detection of micro-calcifications, despite being trained using only image-level labels.
\end{abstract}

\keywords{breast cancer detection, screening mammogram, deep learning, self-supervised learning, swin transformer, self-attention}

{\noindent \footnotesize\textbf{*}Address all correspondence to Han Chen, \linkable{han.chen@sri.utoronto.ca} }

\begin{spacing}{2}   

\section{Introduction}
\label{sect:intro}  
Breast cancer is one of the most commonly diagnosed cancers worldwide and remains the leading cause of cancer-related mortality among women\cite{sung2021global}. Early detection through screening mammograms is one of the most effective strategies to improve survival rates\cite{pashayan2020personalized}. Traditionally, the interpretation of mammograms is performed manually by radiologists, often supplemented by computer-aided diagnosis (CAD) tools to assist in detecting and classifying breast lesions\cite{zebari2021systematic}. Although screening mammography has significantly reduced breast cancer mortality\cite{duffy2021beneficial}, a recent review of mammograms diagnosed by radiologists has revealed limitations, including numerous false negatives\cite{hovda2022true}. To address this issue, double reading has been introduced in some regions to reduce missed cancers\cite{mckinney2020international,yan2023automated}. However, this approach is time-intensive and increases the workload for radiologists. As a result, achieving accurate and efficient early screening for breast cancer remains critical and challenging.

Recent studies \cite{schaffter2020evaluation,lotter2021robust} have highlighted the potential of modern AI systems in mammography, showing that deep learning-enhanced models can perform on par with human readers and often surpass them when integrated with radiologists' decisions. These AI techniques eliminate the need for manual intervention by incorporating feature extraction and selection directly into deep neural networks \cite{samek2021explaining,chen2023teacher,chen2022unsupervised,chen2022pose}. Convolutional neural networks (CNN), initially developed for natural image processing, have been widely adopted for automated mammogram analysis and have demonstrated significant promise in assisting radiologists \cite{shen2021interpretable,ribli2018detecting,rangarajan2022ultra}. However, CNNs face limitations due to their small convolutional kernel sizes, which restrict their receptive field and hinder their ability to capture global contextual information. This limitation is particularly significant for mammogram analysis, where regions of interest (ROIs), such as masses and micro-calcifications, are often much smaller and less conspicuous than objects in natural images \cite{guo2022cmt}. 

Moreover, accurate breast cancer diagnosis relies on both local details, such as lesion shape, and global structures, like overall breast density and tissue patterns \cite{pinto2009spatial,wei2011association,shen2021interpretable}. While CNN-based models have been the dominant architecture in medical imaging, the emergence of Vision Transformers (ViTs), such as the Swin Transformer (Swin-T) \cite{liu2021swin} has brought new possibilities. ViTs excel in modelling long-range dependencies and complex spatial relationships, providing an advantage over CNN. Its self-attention design effectively captures both local and global features, positioning ViTs as a promising framework for advancing automated mammogram analysis and enhancing the overall performance of AI-based breast cancer diagnosis systems.

Deep learning models, especially Transformer based models, are inherently data-hungry and often achieve better performance with larger datasets. However, curating extensive medical imaging datasets is both costly and time-consuming. Unlike photographic images, which can be sourced online and annotated by non-experts, radiographic images require domain expertise for accurate labeling. Self-supervised learning (SSL) addresses this challenge by eliminating the need for manual annotations. SSL trains models to learn generalizable features by solving pretext tasks, such as predicting image rotations or generating contrastive representations\cite{jing2020self}, significantly reducing the reliance on labelled medical data. Recent SSL techniques, such as DINO\cite{caron2021emerging}, EsViT\cite{li2021efficient}, and BYOL\cite{grill2020bootstrap}, have demonstrated remarkable success in pretraining both CNN and ViT models. These methods enable models to extract high-quality feature representations, which can then be fine-tuned for downstream tasks such as medical image classification. In particular, the Efficient Self-supervised Vision Transformers (EsViT) employs a patch-based contrastive learning framework combined with token sparsification, emphasizing the most informative regions within an image. This approach not only reduces computational costs but also preserves the ability to extract high-quality features, effectively balancing fine-grained and global feature learning. Its capacity to handle high-resolution images makes EsViT especially well-suited for analyzing large-scale mammograms, where both localized details and broader contextual information are critical for accurate diagnosis.

To further enhance the potential of deep learning for early breast cancer detection on screening mammograms, we extend our previous work \cite{chen2024towards}, where we introduced a classification network for breast cancer detection. In this extended study, we introduce a novel methodology by incorporating self-supervised pretraining with the HybMNet architecture, improving its performance. The key innovation in this work lies in the integration of SSL pretraining to enhance the Swin-T backbone’s ability to capture complex patterns in mammograms, especially in the context of limited training data.  To the best of our knowledge, we are the first to apply SSL pretraining to high-resolution mammogram inputs, further advancing the potential for more accurate breast cancer detection.

Specifically, we pretrain the Swin-T backbone using the EsViT SSL approach on a small mammogram dataset, initializing it with weights from SSL-pretrained ImageNet-1K models. This two-stage pretraining strategy improves the backbone’s feature extraction capabilities, especially for mammograms, by leveraging knowledge from large-scale image datasets before fine-tuning on domain-specific data. The pretrained Swin-T is then used in the HybMNet architecture, where it identifies key regions in the mammogram through hierarchical feature maps and self-attention within locally shifted windows, enhancing its ability to capture relationships between ROIs and surrounding tissue compared to traditional convolutional methods. These identified ROIs are then processed by a CNN-based network to extract finer-grained features, with a fusion module combining both branches for final predictions. We evaluate the method on publicly available datasets, including CMMD and INbreast, for breast cancer classification task. The results demonstrate that the SSL-pretrained HybMNet achieves improvements in classification performance, including higher AUC, accuracy, and F1-score, compared to our prior work, highlighting the substantial benefits of incorporating SSL pretraining into the network and establishing its potential for advancing breast cancer detection.

\section{Materials and Method}
\begin{figure} [ht]
\begin{center}
\begin{tabular}{c} 
\includegraphics[height=9cm]{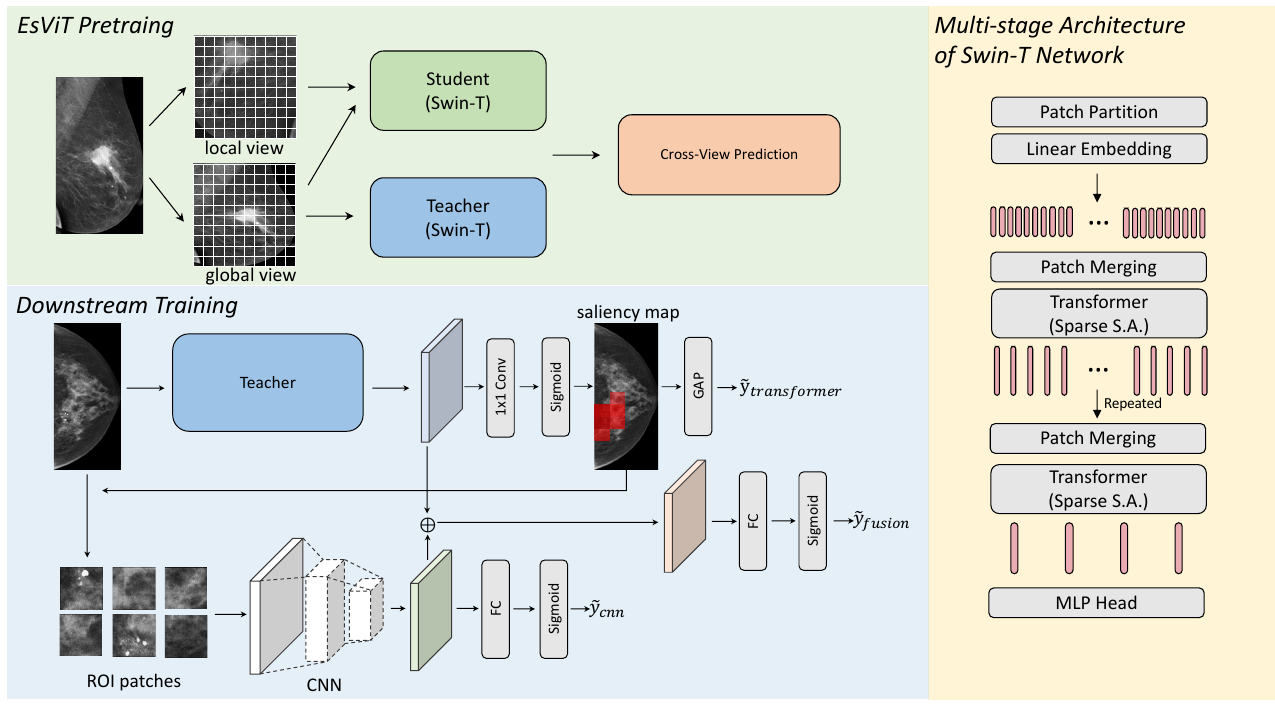}
\end{tabular}
\end{center}
\caption[Overview of the proposed breast cancer detection method] 
{ \label{fig:network_architecture} 
Overview of the proposed breast cancer detection method, consisting of two parts: (1) \textbf{SSL pretraining of the Swin-T backbone using EsViT} on full-resolution mammograms. The multi-stage Swi-T processes each mammogram by organizing it into a sequence of smaller patches. Sparse self-attention mechanisms are applied to capture fine-grained features while reducing computational complexity. At intermediate layers, neighboring tokens are merged hierarchically to further improve efficiency. The pretrained Swin-T model with the lowest loss is then used to initialize the feature extractor for the downstream HybMNet. (2) \textbf{Downstream training of HybMNet} that integrates a Swin-T branch and a CNN branch. The Swin-T branch utilizes the pretrained Swin-T to generate a saliency map, identifying the most informative regions (ROIs) in the input mammogram. These candidate ROI patches are cropped from the original image and passed to the CNN branch for extracting detailed local features. A fusion module combines global features from the Swin-T branch with local features from the CNN branch to produce robust overall predictions for breast cancer detection.}
\end{figure}

We proposed a method for automatically distinguishing between malignant and benign (normal) mammograms, leveraging a SSL technique and the HybMNet that integrates local self-attention with fine-grained feature extraction. The overall pipeline of our proposed approach is illustrated in Fig.\ref{fig:network_architecture}. In the following sections, we provide an overview of the datasets used in our research and a detailed explanation of the key components of our algorithm.

\subsection{Data}
This study includes two public datasets: 
\\
(1) The Chinese Mammogram Database (CMMD)\cite{cai2023online} dataset contains full-field digital mammograms (FFDMs) collected by researchers at South China University of Technology. It contains mammograms from 1,775 patients. The type of benign or malignant tumours was confirmed by biopsy. All images were produced by GE Senographe DS mammography system and a Siemens Mammomat Inspiration mammography system. We obtained this dataset through The Cancer Imaging Archive (TCIA)\cite{clark2013cancer} Public Access. While this dataset includes multiple images from one exam for each patient, we treated each image as a separate sample. To construct pretraining, training and test sets, we performed data splitting at the patient level, ensuring that all images from the same patient are in the same set. The data split for CMMD is detailed in Table \ref{tab:cmmd_split}.

\begin{table}[ht]
\caption{Data split for the CMMD dataset, showing the distribution of images across the pretraining, training, and test sets, with splits performed at the patient level to ensure no overlap between sets.} 
\label{tab:cmmd_split}
\begin{center}
\begin{tabular}{|l|c|c|c|}
\hline
\rule[-1ex]{0pt}{3.5ex} Number & Pretraining & Train & Test \\
\hline
\rule[-1ex]{0pt}{3.5ex}  case-level & 1,289 & 300 & 186 \\
\hline
\rule[-1ex]{0pt}{3.5ex}  image-level & 3,832 & 916 & 526 \\
\hline 
\end{tabular}
\end{center}
\end{table}

\noindent (2) The INbreast dataset \cite{moreira2012inbreast} is a public collection of FFDM images in DICOM format. It consists of 410 mammograms from 115 unique patients, with 107 cases presenting mass lesions in both MLO and CC views. The original mammograms have an average resolution of 3328 × 4084 or 2560 × 3328 pixels and include pixel-level annotations as well as class labels, such as pathology and BI-RADS categories. The acquisition equipment was the MammoNovation Siemens system. Since the INbreast dataset lacks explicit ground truth for benign or malignant labels, we adopted the approach proposed in \cite{shen2017end}, manually assigning BI-RADS readings of 1 and 2 as negative samples and readings of 4, 5, and 6 as positive samples, while excluding images with BI-RADS 3. Consequently, the task on the INbreast dataset effectively becomes a binary classification task. We performed an 80-20 patient-level stratified split into training and test sets, resulting in 308 images in the training set and 76 images in the test set.

\begin{table}[ht]
\caption{Data split for the INbreast dataset, showing the distribution of images across the training and test sets, with splits performed at the patient level to ensure no overlap between sets.} 
\label{tab:inbreast_split}
\begin{center}
\begin{tabular}{|l|c|c|c|}
\hline
\rule[-1ex]{0pt}{3.5ex} Number & Train & Test \\
\hline
\rule[-1ex]{0pt}{3.5ex}  case-level & 85 & 20 \\
\hline
\rule[-1ex]{0pt}{3.5ex}  image-level & 308 & 76 \\
\hline 
\end{tabular}
\end{center}
\end{table}

\subsection{Self-supervised Pretraining via EsViT}
Self-supervised pretraining enables the learning of general-purpose visual representations without relying on manual annotations, making it particularly effective for medical image analysis, where labelled data is often scarce. In this work, we use the Swin-T Base \cite{liu2021swin} as our backbone and employ the EsViT (Efficient Self-Supervised Vision Transformers) method \cite{li2021efficient} for pretraining. In addition, we introduce several adaptations to EsViT to optimize it for screening mammograms.

\sethlcolor{yellow}
\subsubsection{\texorpdfstring{Structure of EsViT}{Structure of EsViT}}
EsViT employs a non-contrastive self-supervised learning framework based on the knowledge distillation paradigm from DINO (Self-Distillation with No Labels) \cite{caron2021emerging}. In this setup, a student network is trained to match the output of a teacher network, both of which consist of a Transformer backbone and a projection head. The student learns from augmented views of an image, generating feature maps that are processed through two MLP heads: one for view-level predictions and the other for region-level predictions. EsViT’s hierarchical transformer structure efficiently handles high-resolution images by dividing them into multi-scale patches, capturing both fine-grained local details and long-range dependencies. It uses relative position bias to ensure consistent spatial relationships across varying cropping resolutions. In the teacher-student framework, the student is updated via backpropagation, while the teacher model is updated slowly using a momentum mechanism to provide stable supervision. Additionally, EsViT integrates region-level tasks to capture inter-region relationships, further improving its ability to learn fine-grained spatial dependencies.

\subsubsection{Adapting EsViT for mammogram pretraining}
We pretrain the Swin-T Base model using EsViT on whole mammograms, considering that mammograms are high-resolution images where fine-grained details are crucial for accurate predictions. Current SSL methods are primarily designed for natural images with input resolutions no larger than \(384 \times 384\), which limits their direct applicability to mammograms. To optimize EsViT for this task, we made the following adaptations:

(1) Preprocessing: We identified the largest connected component of non-zero pixels in each mammogram and cropped the rectangular region containing it. This ensures the focus remains on the relevant areas instead of background.

(2) Input Resizing: Instead of using the random resized crop strategy originally employed in EsViT, we resized the global inputs to \(512 \times 512\) and the local inputs to the same size before applying a random resized crop (\(300 \times 300\)) to the local inputs. This adjustment was made to better capture key areas such as masses or micro-calcifications, which are often highly localized.

(3) Window Size Adjustment: Following research \cite{ren2022beyond} demonstrating that larger window sizes improve the performance of Swin-T, we increased the default window size from 7 to 16 during pretraining to enhance the model's ability to capture long-range dependencies in high-resolution mammograms.

These modifications allow EsViT to better leverage the unique characteristics of mammograms, improving its ability to extract meaningful features for downstream tasks.

\subsection{Downstream Training of HybMNet}
The architecture of HybMNet is shown in Fig.\ref{fig:network_architecture}. It is a hybrid model designed for mammogram classification, integrating both global and local feature extraction techniques. The model utilizes the pretrained Swin-T backbone to extract global features through its hierarchical transformer structure. To enhance performance further, a CNN-based module is employed to extract local detailed features from ROIs identified by a saliency map. Both the Swin-T and CNN branches provide individual predictions, and the features from these two branches are also fused to calculate the final overall prediction. All predictions are then used for loss calculation.

\subsubsection{Global feature extraction via Swin-T}
After pretraining the Swin-T on full mammograms, it is used as the backbone for the downstream classification network. The weights achieving the lowest loss during pretraining are selected for initialization in the downstream training. The choice of Swin-T as the backbone is driven by its demonstrated ability to capture global features through the hierarchical organization of feature maps, which has been shown to outperform its predecessor, the Vision Transformer, in both accuracy and efficiency for X-ray analysis \cite{ma2022benchmarking}. 

Our feature extractor leverages the Swin-T's local self-attention mechanism, partitioning the high-resolution mammogram (\(1024 \times 1024\)) into small windows (\(16 \times 16\), with each patch measuring \(4 \times 4\)), and performing attention operations within each window simultaneously. By gradually merging neighboring windows, the model generates pyramid-shaped feature maps with progressively reduced spatial dimensions and increased depth. This hierarchical approach enables the Swin-T to effectively handle high-resolution inputs while preserving essential details. Unlike conventional methods that rely on aggressive downsampling to process mammograms via ViTs, our backbone leverages Swin-T's local self-attention to capture meaningful contextual information from a global perspective without sacrificing spatial resolution.

\subsubsection{Saliency map and ROI patches generation}
To extract candidate ROIs, we incorporate a \(1 \times 1\) convolutional layer combined with a sigmoid activation function to convert the output feature representations from the backbone into a saliency map. In this saliency map, each value represents the probability of the corresponding location contributing to the final classification result, effectively highlighting regions where masses or micro-calcifications are most likely to occur. It is then sent to global average pooling (GAP) to obtain the global prediction \(\hat{y}_{\text{Swin-T}}\).

Based on the saliency map, we also generate 6 ROI patches using a greedy algorithm inspired by \cite{shen2021interpretable}, it is achieved by maximizing the sum of saliency values within each patch. This approach identifies regions most likely to contain important features. The position of each selected patch is recorded, and the corresponding saliency map values are set to zero to prevent re-selection. The selected patches are then cropped from the original high-resolution mammograms and used as input for the CNN, where fine-grained feature extraction is performed for further enhancing the classification process.

\subsubsection{Fine-grained local feature extraction from ROI patches via CNN}
To effectively capture the subtle local details present in mammograms, we utilize a CNN to extract feature representations from the candidate ROI patches. Specifically, we employ a ResNet-18 \cite{he2016deep}, which consists of 18 layers organized into four stacks of residual blocks. Each block comprises two convolutional layers, with the output feature map of one layer serving as the input for the next. 

The CNN processes the candidate patches independently, extracting fine-grained features from each. The feature maps generated for all candidate patches are concatenated and passed through a fully connected layer with a sigmoid activation function to produce a prediction \(\hat{y}_{\text{CNN}}\). Additionally, these CNN-extracted feature maps are integrated with the global features from the Swin-T backbone through a fusion module, generating the final combined prediction. This dual-path approach ensures that both local and global information are effectively captured and leveraged for robust breast cancer detection.

\subsubsection{Fusion of global and local feature representations}
To integrate information from both the Swin-T and CNN branches, we concatenate the feature maps extracted by the Swin-T with those from the CNN. The combined representation is then passed through a fully connected layer with a sigmoid activation function to produce the final prediction. This is analogous to modeling the diagnosis process of a radiologist considering the global and local information to deliver a comprehensive result.

\subsubsection{Loss function}
Our proposed network is trained in an end-to-end manner, where the overall learning objective encompasses the predictions from the Swin-T branch, the CNN branch, and the fusion part. The loss function is defined as follows,

\begin{equation}
\label{eq:fov}
\theta^{*} = \underset{\theta}{\arg \min} \, \Big( L_{\text{bce}}(y, \hat{y}_{\text{Swin-T}}) + L_{\text{bce}}(y, \hat{y}_{\text{CNN}}) + L_{\text{bce}}(y, \hat{y}_{\text{fusion}}) + \alpha \sum_{(i,j)} \left| \text{saliency\_map}_{(i,j)} \right| \Big)
\end{equation}

\noindent where \( L_{bce}(\cdot) \) represents the binary cross-entropy loss, and \( y \) denotes the classification label, indicating whether the input is benign or malignant. The indices \( (i,j) \) refer to the pixel locations in the saliency map. Additionally, we apply \( L1 \) regularization to the saliency map to promote sparsity and highlight the most informative regions, with \( \alpha \) serving as a hyperparameter. The contribution of each component in our loss function is further examined in the ablation study section \ref{ablation_study}.

\section{Results}
\label{sect:sections}

\subsection{Implementation Details}
(1) SSL pretraining: We use Swin-T Base network as backbone and pretrained it on the CMMD pretraining set using EsViT, with initialization from weights derived through self-supervised pretraining on the ImageNet-1K dataset. This double pretraining strategy effectively combines the general feature representations from ImageNet-1K with the domain-specific knowledge of the CMMD dataset to address the limited availability of mammogram data. To accommodate a larger window size (increased from 7 to 16), bicubic interpolation was applied to adapt the pretrained model weights. We follow the original data augmentation settings of EsViT, including flipping, ColorJitter, GaussianBlur, and Solarization, to generate varied image views and enhance the learning of robust features.

The pretraining was conducted on two NVIDIA A100 GPUs with a batch size of 4 per GPU. An initial learning rate of 0.0005 was employed and decayed to a minimum of \(1 \times 10^{-6}\) using a cosine annealing schedule. AdamW \cite{loshchilov2017decoupled} served as the optimizer, with a weight decay of 0.04. The teacher-student framework maintained a teacher momentum of 0.996 and employed a softmax temperature of 0.07 to regulate prediction sharpness. Training included a 10-epoch warmup phase for stability, followed by a total of 30 epochs to ensure convergence of the SSL objectives. 

(2) Downstream training: We begin by selecting the largest connected component in each input mammogram and cropping the breast region. The cropped region is then resized to \(1024 \times 1024\) pixels. To enhance training generalization, we apply data augmentation techniques, including random vertical flipping and random affine transformations. Additionally, all mammograms are reoriented to ensure the breast consistently appears on the left side of the image.

The model is optimized using the AdamW optimizer with a learning rate of \(5 \times 10^{-5}\), weight decay of 0.05, a batch size of 8, and a total of 200 training epochs. The learning rate is warmed up for the first 10 epochs, starting at \(2.5 \times 10^{-7}\), before transitioning to the target value. The hyperparameter \(\alpha\) is set to \(3.26 \times 10^{-6}\).

All experiments are implemented in PyTorch \cite{paszke2019pytorch} and conducted on an NVIDIA A100 GPU. Model performance is evaluated using the Area Under the ROC Curve (AUC), accuracy, and F1-score, the threshold for calculating accuracy and F1-score is 0.5. For experiments on the CMMD dataset, 80\% of the test set is randomly sampled, and the process is repeated five times to calculate the 95\% confidence intervals (CIs) for each metric. For the INbreast dataset, a 5-fold cross-validation is performed, and 95\% CIs are calculated for all evaluation metrics.

\subsection{Comparison with State-of-the-art Methods}
We compare our proposed method with GMIC\cite{shen2021interpretable}, a previous study designed for breast cancer detection on digital mammograms. We fine-tuned GMIC in two different ways: (1) GMIC$_{fc}$, where we load the model checkpoint pretrained on approximately 0.8M mammographic images from the NYU training set to initialize the feature extractor backbone for GMIC and freeze it during training. The classifier is then fine-tuned using our training sets. (2) GMIC, where we initialize the entire model with weights pretrained on the NYU training set and fine-tune the entire model using our training data.

\begin{figure} [ht]
\begin{center}
\begin{tabular}{c} 
\includegraphics[height=8cm]{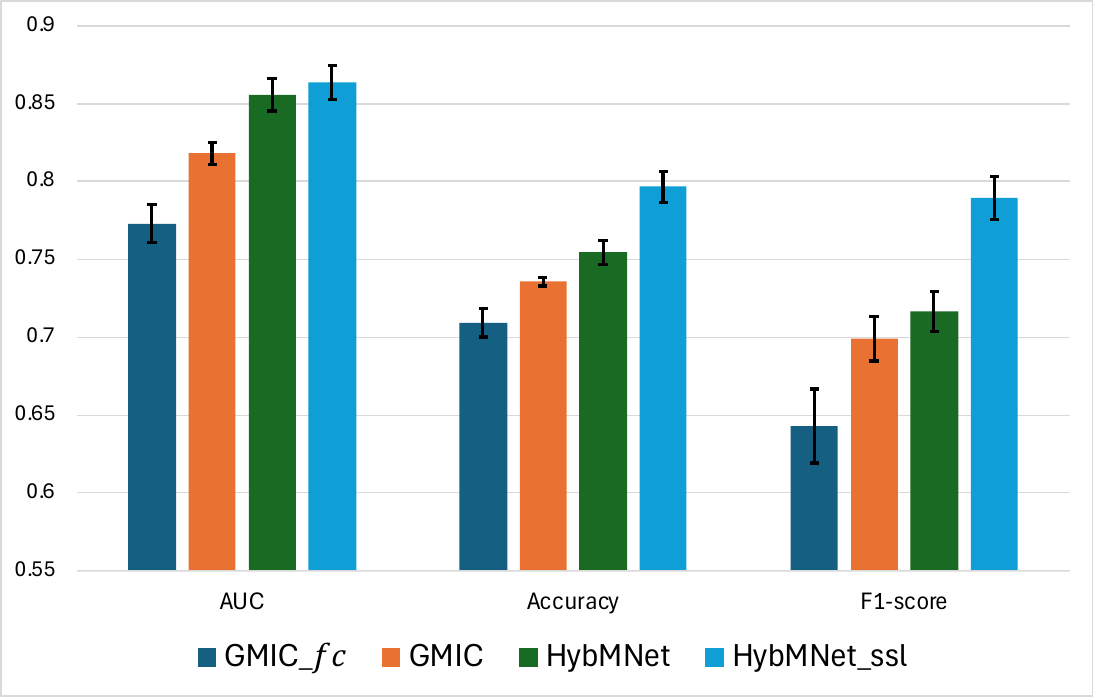}
\end{tabular}
\end{center}
\caption[Comparison with state-of-the-art methods on CMMD test set] 
{ \label{fig:sota_CIs_cmmd}
Comparison with state-of-the-art methods on CMMD test set. Error bars represent 95\% CIs and the centers correspond to the mean of each classification metric across different models.
}
\end{figure}

Table \ref{tab:classification-results_cmmd} presents the classification results on the CMMD test set. The learning objective is to distinguish whether the input mammogram shows a malignant lesion. Thus during both training and testing, mammograms from patients diagnosed with normal and benign conditions were treated as the same class. Our model outperforms the state-of-the-art GMIC model, despite being initialized with out-of-domain ImageNet-pretrained weights and trained on a small-scale dataset. It exceeds the performance of the compared method, which was trained on a large mammogram database, across all three evaluation metrics. When SSL pretraining with the unlabeled CMMD data is incorporated, performance improves further, particularly in the F1-score (rising from 0.717 (95\% CI: 0.703, 0.729) with ImageNet-pretrained weights to 0.790 (95\% CI: 0.776, 0.803)). The statistical significance of the improvement between HybMNet and HybMNet$_{ssl}$ is further supported by a paired t-test for AUC, which yielded a p-value of 0.048, indicating that the improvement is statistically significant at the 5\% significance level. These results highlight the effectiveness of our proposed network for breast cancer detection, even with relatively small datasets. The error bars shown in Fig.\ref{fig:sota_CIs_cmmd} further demonstrate that our method achieves competitive results across all three metrics, with a notably more stable F1-score, reflecting a strong balance between precision and recall.

\begin{table}[ht]
\caption{Benign vs.Malignant classification results on CMMD test set. The values in parentheses represent the 95\% CIs.} 
\label{tab:classification-results_cmmd}
\begin{center} 
\begin{tabular}{|l|c|c|c|}
\hline
\rule[-1ex]{0pt}{3.5ex}  Methods & AUC & Accuracy & F1-score  \\
\hline
\rule[-1ex]{0pt}{3.5ex}  GMIC$_{fc}$ & 0.773 (0.761, 0.785) & 0.710 (0.701, 0.719) & 0.643 (0.619, 0.667)   \\
\hline
\rule[-1ex]{0pt}{3.5ex}  GMIC & 0.819 (0.811, 0.826) & 0.736 (0.733, 0.739) & 0.699 (0.685, 0.714) \\
\hline
\rule[-1ex]{0pt}{3.5ex}  HybMNet & 0.856 (0.846, 0.866) & 0.755 (0.747, 0.762) & 0.717 (0.703, 0.729) \\
\hline 
\rule[-1ex]{0pt}{3.5ex}  HybMNet$_{ssl}$ & 0.864 (0.852, 0.875) & 0.797 (0.787, 0.807) & 0.790 (0.776, 0.803) \\
\hline 
\end{tabular}
\end{center}
\end{table}

Table \ref{tab:classification-results_inbreast} presents the results on the INbreast test set, where our proposed method outperforms the compared approaches in both AUC and F1-score. To evaluate the generalizability of our SSL-pretrained backbone for mammograms, we initialized the Swin-T model with weights pretrained on both ImageNet and CMMD datasets for the downstream experiments with INbreast.  The paired t-test between HybMNet and HybMNet$_{ssl}$ for AUC yielded a p-value of 0.32, indicating no statistically significant improvement on this particular dataset, this may be attributed to differences in intensity distributions between the pretraining (CMMD) and downstream (INbreast) datasets. However, the marginal increase in AUC achieved by HybMNet${ssl}$ still demonstrates that SSL pretraining in our approach can transfer across datasets. Additionally, the observed increase in Accuracy and F1-score further supports the robustness of our method.

\begin{table}[ht]
\caption{Binary classification} results on INbreast test set. The values in parentheses represent the 95\% CIs.
\label{tab:classification-results_inbreast}
\begin{center} 
\begin{tabular}{|l|c|c|c|}
\hline
\rule[-1ex]{0pt}{3.5ex}  Methods & AUC & Accuracy & F1-score  \\
\hline
\rule[-1ex]{0pt}{3.5ex}  GMIC$_{fc}$ & 0.758 (0.734, 0.781) & 0.700 (0.682, 0.718) & 0.545 (0.522, 0.569) \\
\hline
\rule[-1ex]{0pt}{3.5ex}  GMIC & 0.843 (0.821, 0.865) & 0.783 (0.761, 0.806) & 0.623 (0.581, 0.666) \\
\hline
\rule[-1ex]{0pt}{3.5ex}  HybMNet & 0.877 (0.848, 0.907) & 0.753 (0.727, 0.780) & 0.678 (0.646, 0.709) \\
\hline 
\rule[-1ex]{0pt}{3.5ex}  HybMNet$_{ssl}$ & 0.889 (0.875, 0.903) & 0.830 (0.811, 0.849) & 0.719 (0.690, 0.747) \\
\hline 
\end{tabular}
\end{center}
\end{table}

\begin{figure} [ht]
   \begin{center}
   \begin{tabular}{c} 
    \includegraphics[height=8cm]{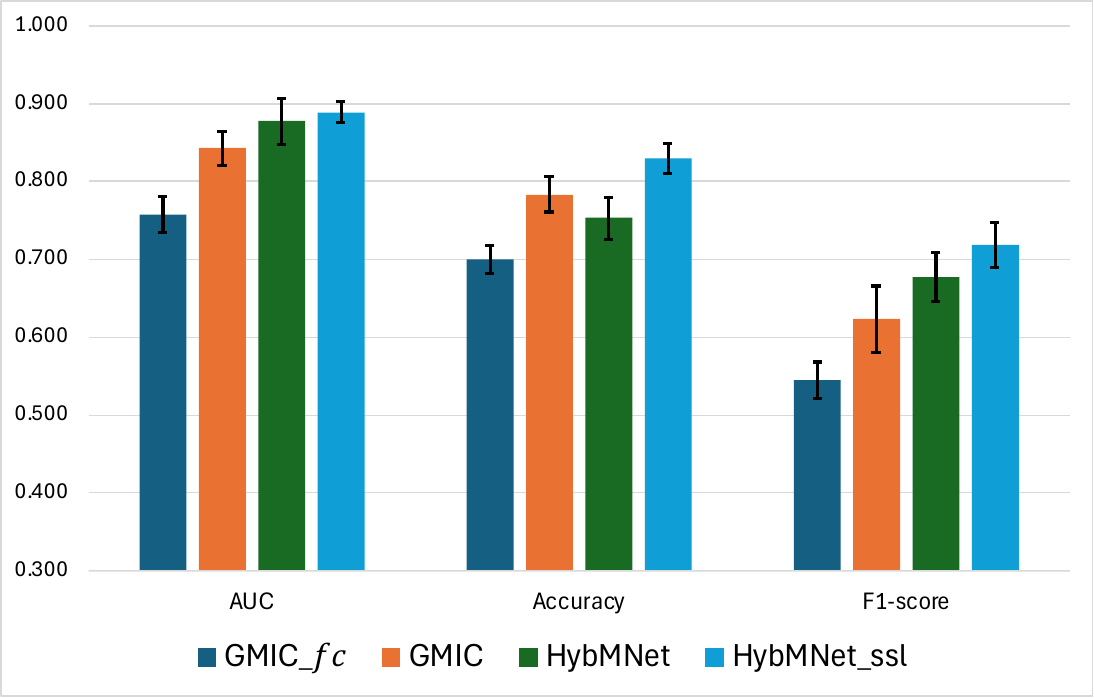}
	\end{tabular}
	\end{center}
   \caption[Comparison with state-of-the-art methods on INbreast test set] 
   { \label{fig:sota_CIs_inbreast}
   Comparison with state-of-the-art methods on INbreast test set. Error bars represent 95\% CIs and the centers correspond to the mean of each classification metric across different models.
   }
\end{figure}

\subsection{Ablation Study}
\label{ablation_study}
\begin{figure} [ht]
   \begin{center}
   \begin{tabular}{c} 
    \includegraphics[height=8cm]{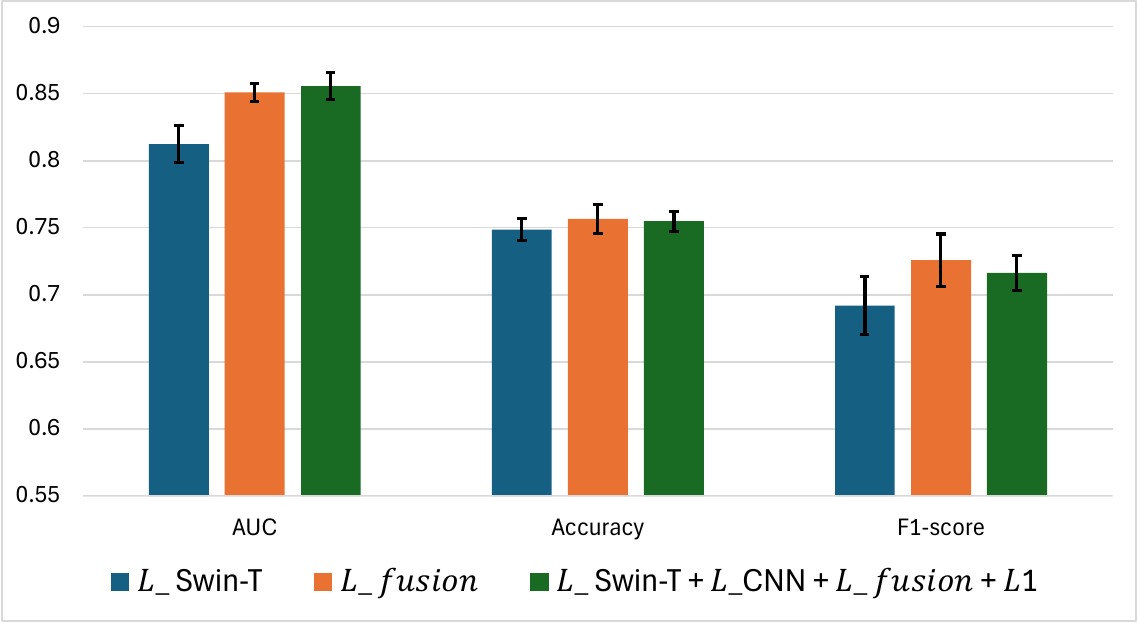}
	\end{tabular}
	\end{center}
   \caption[Ablation study results on CMMD test set] 
   { \label{fig:ablation_CIs}
   Ablation study results on CMMD test set. Error bars represent 95\% CIs and the centers correspond to the mean of each classification metric across different models.
   }
\end{figure}

To evaluate the contribution of each component of our model, we conducted an ablation study, with the results summarized in Table \ref{tab:ablation-study}. All experiments were initialized with ImageNet-1k pretrained weights. Starting with the Swin-T backbone, the model achieved an mean AUC of 0.812. When representations extracted by the backbone and the CNN model were combined and fused for the final prediction, improvements were observed across all three evaluation metrics (as shown in the second row). The best performance was achieved by integrating predictions from the Swin-T, CNN, and fusion modules, along with incorporating $L1$ regularization on the saliency map. Additionally, we visualized the results with error bars in Fig. \ref{fig:ablation_CIs}, which highlight that integrating both global and local information produces more stable predictions, as evidenced by narrower CIs. These findings demonstrate the effectiveness of our proposed architecture in leveraging meaningful features from both global and local contexts.

\begin{table}[ht]
\caption{Ablation study of contribution of each component in our proposed network. Data in parentheses are 95\% CIs.} 
\label{tab:ablation-study}
\begin{center}
\resizebox{\textwidth}{12mm}{
\begin{tabular}{|l|c|c|c|}
\hline
\rule[-1ex]{0pt}{3.5ex}  Methods & AUC & Accuracy & F1-score \\
\hline
\rule[-1ex]{0pt}{3.5ex}  ${L}_{Swin-T}$ & 0.812 (0.799, 0.826) & 0.749 (0.740, 0.757) & 0.692 (0.670, 0.714) \\
\hline
\rule[-1ex]{0pt}{3.5ex}  ${L}_{fusion}$ & 0.851 (0.844, 0.858) & 0.757 (0.746, 0.767) & 0.726 (0.707, 0.745) \\
\hline
\rule[-1ex]{0pt}{3.5ex}  ${L}_{fusion}$ +  ${L}_{Swin-T}$ +  ${L}_{CNN}$ + ${L1}$ & 0.856 
(0.846, 0.866) & 0.755 (0.747, 0.762) & 0.717 (0.703, 0.729) \\
\hline 
\end{tabular}
}
\end{center}
\end{table}

\subsection{Visualization of ROI patches}
\begin{figure} [ht]
   \begin{center}
   \begin{tabular}{c} 
   \includegraphics[height=6cm]{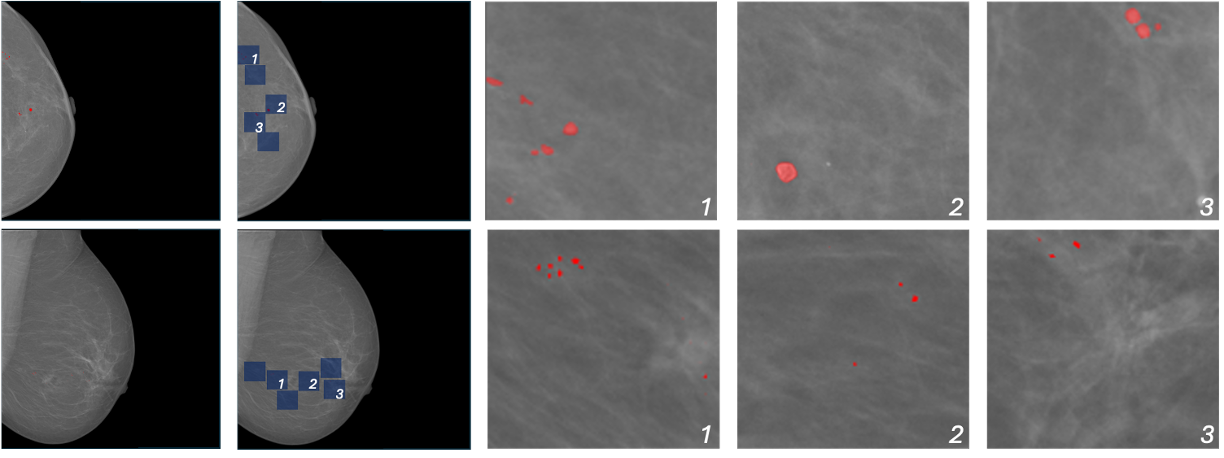}
	\end{tabular}
	\end{center}
   \caption[Visualization of ROI patches] 
   { \label{fig:calcification_detection}
    Visualization of ROI patches. From left to right, the input mammograms annotated with calcification segmentation labels, the identified ROI patch locations (highlighted in blue), and three selected ROI patches containing calcifications (marked in red).
   }
\end{figure}

To highlight the effectiveness of HybMNet, we visualize several ROI patches generated from the saliency maps in Fig. \ref{fig:calcification_detection}. Both the input mammograms and their corresponding ground truth segmentation maps are derived from the INbreast dataset. The qualitative results demonstrate that, even without pixel-level annotations during training, our network effectively leverages image-level labels to automatically recognize the most informative regions contributing to the final predictions. Notably, in the second example, the model can identify  micro-calcifications that are difficult to detect with the human eye, underscoring its potential for other tasks such as outcome prediction, where more subtle changes in appearance may play a critical role.

\section{Discussion}
\textbf{Impact of Domain Shift.} We proposed a two-stage learning process for breast cancer detection, incorporating self-supervised pretraining on CMMD followed by downstream training and testing on CMMD and INbreast datasets. Our in-domain experiments demonstrated that HybMNet$_{ssl}$, initialized with self-supervised learning, achieved a statistically significant improvement in AUC compared to HybMNet trained only with ImageNet-1K pretraining. This highlights the effectiveness of SSL in capturing meaningful representations when the pretraining and downstream data share a similar domain.

In the out-of-domain setting, where the downstream dataset (INbreast) differs from the pretraining dataset (CMMD) in imaging characteristics, HybMNet$_{ssl}$ showed only a marginal AUC improvement over HybMNet. This is likely due to domain discrepancies caused by differences in imaging equipment and protocols. Prior study \cite{rani2024self} has shown that SSL pretraining helps models adapt learned representations to related but unseen data, improving performance and generalization. Therefore, despite similar results on INbreast, HybMNet$_{ssl}$ is expected to have better generalization and robustness.

\noindent\textbf{Enhancing Domain Generalization.} While this study employed the standard augmentation method from EsViT during SSL pretraining, future research will explore more advanced augmentation strategies, such as intensity transformations and gamma correction, to improve feature invariance and generalization. To further improve the generalization of our approach, future work will incorporate a more diverse set of mammography data collected from different manufacturers and imaging protocols. This will help mitigate domain shift issues and enhance model robustness across varied clinical settings.

\section{Conclusions}
We introduced a novel approach for high-resolution mammogram classification, starting with SSL pretraining to enhance feature extraction capabilities. Specifically, the Swin-T backbone was pretrained using EsViT on mammograms, with initialization from SSL-pretrained ImageNet-1K weights. This double pretraining strategy effectively addresses the challenge of limited mammogram data and improves the network's ability to capture meaningful representations. Building on this foundation, we proposed a HybMNet that combines the Swin-T and CNN to model both global features from the entire mammogram and fine-grained local details within ROI patches. Evaluations on public datasets, including CMMD and INbreast, demonstrate the superiority of our method, achieving state-of-the-art performance in distinguishing between benign and malignant cases. This approach holds promise for advancing CAD systems, aiding radiologists in prioritizing suspicious cases or serving as a reliable second reader in clinical workflows.

\section{Disclosures}
The authors have no relevant financial interests in the manuscript and no other potential conflicts of interest to disclose.

\section{Code and Data Availability}
The code used for this study is not publicly accessible but may be provided to qualified researchers upon reasonable request to the corresponding author. The datasets used for this study, including CMMD and INbreast, are publicly available. The CMMD dataset can be accessed at https://www.cancerimagingarchive.net/collection/cmmd/, and the INbreast dataset can be found at https://www.kaggle.com/datasets/tommyngx/inbreast2012.

\section{Acknowledgments}
This work was funded by the Canadian Institutes of Health Research, the Canada Foundation for Innovation (Grant No. 40206), and the Ontario Research Fund.


\bibliography{report}   

\begin{thebibliography}{10}

\bibitem{sung2021global}
H.~Sung, J.~Ferlay, R.~L. Siegel, {\em et~al.}, ``Global cancer statistics 2020: Globocan estimates of incidence and mortality worldwide for 36 cancers in 185 countries,'' {\em CA: a cancer journal for clinicians} {\bf 71}(3), 209--249  (2021).

\bibitem{pashayan2020personalized}
N.~Pashayan, A.~C. Antoniou, U.~Ivanus, {\em et~al.}, ``Personalized early detection and prevention of breast cancer: Envision consensus statement,'' {\em Nature Reviews Clinical Oncology} {\bf 17}(11), 687--705  (2020).

\bibitem{zebari2021systematic}
D.~A. Zebari, D.~A. Ibrahim, D.~Q. Zeebaree, {\em et~al.}, ``Systematic review of computing approaches for breast cancer detection based computer aided diagnosis using mammogram images,'' {\em Applied Artificial Intelligence} {\bf 35}(15), 2157--2203  (2021).

\bibitem{duffy2021beneficial}
S.~W. Duffy, L.~Tab{\'a}r, A.~M.-F. Yen, {\em et~al.}, ``Beneficial effect of consecutive screening mammography examinations on mortality from breast cancer: a prospective study,'' {\em Radiology} {\bf 299}(3), 541--547  (2021).

\bibitem{hovda2022true}
T.~Hovda, S.~R. Hoff, M.~Larsen, {\em et~al.}, ``True and missed interval cancer in organized mammographic screening: a retrospective review study of diagnostic and prior screening mammograms,'' {\em Academic Radiology} {\bf 29}, S180--S191  (2022).

\bibitem{mckinney2020international}
S.~M. McKinney, M.~Sieniek, V.~Godbole, {\em et~al.}, ``International evaluation of an ai system for breast cancer screening,'' {\em Nature} {\bf 577}(7788), 89--94  (2020).

\bibitem{yan2023automated}
F.~Yan, H.~Huang, W.~Pedrycz, {\em et~al.}, ``Automated breast cancer detection in mammography using ensemble classifier and feature weighting algorithms,'' {\em Expert Systems with Applications} {\bf 227}, 120282  (2023).

\bibitem{schaffter2020evaluation}
T.~Schaffter, D.~S. Buist, C.~I. Lee, {\em et~al.}, ``Evaluation of combined artificial intelligence and radiologist assessment to interpret screening mammograms,'' {\em JAMA network open} {\bf 3}(3), e200265--e200265  (2020).

\bibitem{lotter2021robust}
W.~Lotter, A.~R. Diab, B.~Haslam, {\em et~al.}, ``Robust breast cancer detection in mammography and digital breast tomosynthesis using an annotation-efficient deep learning approach,'' {\em Nature Medicine} {\bf 27}(2), 244--249  (2021).

\bibitem{samek2021explaining}
W.~Samek, G.~Montavon, S.~Lapuschkin, {\em et~al.}, ``Explaining deep neural networks and beyond: A review of methods and applications,'' {\em Proceedings of the IEEE} {\bf 109}(3), 247--278  (2021).

\bibitem{chen2023teacher}
H.~Chen, Y.~Jiang, H.~Ko, {\em et~al.}, ``A teacher--student framework with fourier transform augmentation for covid-19 infection segmentation in ct images,'' {\em Biomedical Signal Processing and Control} {\bf 79}, 104250  (2023).

\bibitem{chen2022unsupervised}
H.~Chen, Y.~Jiang, M.~Loew, {\em et~al.}, ``Unsupervised domain adaptation based covid-19 ct infection segmentation network,'' {\em Applied Intelligence} {\bf 52}(6), 6340--6353  (2022).

\bibitem{chen2022pose}
H.~Chen, Y.~Jiang, and H.~Ko, ``Pose-guided graph convolutional networks for skeleton-based action recognition,'' {\em IEEE Access} {\bf 10}, 111725--111731  (2022).

\bibitem{shen2021interpretable}
Y.~Shen, N.~Wu, J.~Phang, {\em et~al.}, ``An interpretable classifier for high-resolution breast cancer screening images utilizing weakly supervised localization,'' {\em Medical image analysis} {\bf 68}, 101908  (2021).

\bibitem{ribli2018detecting}
D.~Ribli, A.~Horv{\'a}th, Z.~Unger, {\em et~al.}, ``Detecting and classifying lesions in mammograms with deep learning,'' {\em Scientific reports} {\bf 8}(1), 4165  (2018).

\bibitem{rangarajan2022ultra}
K.~Rangarajan, A.~Gupta, S.~Dasgupta, {\em et~al.}, ``Ultra-high resolution, multi-scale, context-aware approach for detection of small cancers on mammography,'' {\em Scientific reports} {\bf 12}(1), 11622  (2022).

\bibitem{guo2022cmt}
J.~Guo, K.~Han, H.~Wu, {\em et~al.}, ``Cmt: Convolutional neural networks meet vision transformers,'' in {\em Proceedings of the IEEE/CVF Conference on Computer Vision and Pattern Recognition},  12175--12185  (2022).

\bibitem{pinto2009spatial}
S.~M. Pinto~Pereira, V.~A. McCormack, S.~M. Moss, {\em et~al.}, ``The spatial distribution of radiodense breast tissue: a longitudinal study,'' {\em Breast Cancer Research} {\bf 11}(3), 1--12  (2009).

\bibitem{wei2011association}
J.~Wei, H.-P. Chan, Y.-T. Wu, {\em et~al.}, ``Association of computerized mammographic parenchymal pattern measure with breast cancer risk: a pilot case-control study,'' {\em Radiology} {\bf 260}(1), 42--49  (2011).

\bibitem{liu2021swin}
Z.~Liu, Y.~Lin, Y.~Cao, {\em et~al.}, ``Swin transformer: Hierarchical vision transformer using shifted windows,'' in {\em Proceedings of the IEEE/CVF international conference on computer vision},  10012--10022  (2021).

\bibitem{jing2020self}
L.~Jing and Y.~Tian, ``Self-supervised visual feature learning with deep neural networks: A survey,'' {\em IEEE transactions on pattern analysis and machine intelligence} {\bf 43}(11), 4037--4058  (2020).

\bibitem{caron2021emerging}
M.~Caron, H.~Touvron, I.~Misra, {\em et~al.}, ``Emerging properties in self-supervised vision transformers,'' in {\em Proceedings of the IEEE/CVF international conference on computer vision},  9650--9660  (2021).

\bibitem{li2021efficient}
C.~Li, J.~Yang, P.~Zhang, {\em et~al.}, ``Efficient self-supervised vision transformers for representation learning,'' {\em arXiv preprint arXiv:2106.09785} {\bf 0}  (2021).

\bibitem{grill2020bootstrap}
J.-B. Grill, F.~Strub, F.~Altch{\'e}, {\em et~al.}, ``Bootstrap your own latent-a new approach to self-supervised learning,'' {\em Advances in neural information processing systems} {\bf 33}, 21271--21284  (2020).

\bibitem{chen2024towards}
H.~Chen and A.~L. Martel, ``Towards improved breast cancer detection on digital mammograms using local self-attention-based transformer,'' in {\em 17th International Workshop on Breast Imaging (IWBI 2024)},   {\bf 13174}, 455--461, SPIE  (2024).

\bibitem{cai2023online}
H.~Cai, J.~Wang, T.~Dan, {\em et~al.}, ``An online mammography database with biopsy confirmed types,'' {\em Scientific Data} {\bf 10}(1), 123  (2023).

\bibitem{clark2013cancer}
K.~Clark, B.~Vendt, K.~Smith, {\em et~al.}, ``The cancer imaging archive (tcia): maintaining and operating a public information repository,'' {\em Journal of digital imaging} {\bf 26}, 1045--1057  (2013).

\bibitem{moreira2012inbreast}
I.~C. Moreira, I.~Amaral, I.~Domingues, {\em et~al.}, ``Inbreast: toward a full-field digital mammographic database,'' {\em Academic radiology} {\bf 19}(2), 236--248  (2012).

\bibitem{shen2017end}
L.~Shen, ``End-to-end training for whole image breast cancer diagnosis using an all convolutional design,'' {\em arXiv preprint arXiv:1711.05775} {\bf 0}  (2017).

\bibitem{ren2022beyond}
P.~Ren, C.~Li, G.~Wang, {\em et~al.}, ``Beyond fixation: Dynamic window visual transformer,'' in {\em Proceedings of the IEEE/CVF Conference on Computer Vision and Pattern Recognition},  11987--11997  (2022).

\bibitem{ma2022benchmarking}
D.~Ma, M.~R. Hosseinzadeh~Taher, J.~Pang, {\em et~al.}, ``Benchmarking and boosting transformers for medical image classification,'' in {\em MICCAI Workshop on Domain Adaptation and Representation Transfer},  12--22, Springer  (2022).

\bibitem{he2016deep}
K.~He, X.~Zhang, S.~Ren, {\em et~al.}, ``Deep residual learning for image recognition,'' in {\em Proceedings of the IEEE conference on computer vision and pattern recognition},  770--778  (2016).

\bibitem{loshchilov2017decoupled}
I.~Loshchilov and F.~Hutter, ``Decoupled weight decay regularization,'' {\em arXiv preprint arXiv:1711.05101} {\bf 1}  (2017).

\bibitem{paszke2019pytorch}
A.~Paszke, S.~Gross, F.~Massa, {\em et~al.}, ``Pytorch: An imperative style, high-performance deep learning library,'' {\em Advances in neural information processing systems} {\bf 32}  (2019).

\bibitem{rani2024self}
V.~Rani, M.~Kumar, A.~Gupta, {\em et~al.}, ``Self-supervised learning for medical image analysis: a comprehensive review,'' {\em Evolving Systems} {\bf 15}(4), 1607--1633  (2024).

\end{thebibliography}
\bibliographystyle{spiejour}   


\vspace{2ex}\noindent\textbf{Han Chen}, PhD, is currently a postdoctoral fellow at Sunnybrook Research Institute, University of Toronto, Canada. She earned her PhD in Electrical and Computer Engineering from Korea University in 2023. Her research focuses on medical image analysis using machine learning and deep learning. Her work aims to enhance clinical decision-making and automated analysis by leveraging limited labeled data and advancing technologies for precise cancer detection.

\vspace{2ex}\noindent\textbf{Anne L. Martel}, PhD, is a senior scientist at Sunnybrook Research Institute and Professor in the Department of Medical Biophysics at the University of Toronto. Her research focuses on developing artificial intelligence and machine learning techniques for medical imaging, with applications in cancer diagnosis and treatment planning. She leads interdisciplinary teams to advance precision medicine and has made significant contributions to image analysis, quantitative imaging biomarkers, and AI-driven healthcare solutions.


\listoffigures

\end{spacing}
\end{document}